\documentclass[conference, 10pt, a4paper]{IEEEtran}
\usepackage[english]{babel}
\usepackage[utf8x]{inputenc}
\usepackage[T1]{fontenc}

\usepackage{geometry}
\IEEEoverridecommandlockouts    
\usepackage{silence}
\usepackage{psfrag}
\usepackage{amsmath} 
\usepackage{amssymb}
\usepackage{amsfonts}
\usepackage{bbm}
\usepackage{nicefrac}
\usepackage{mathtools}
\usepackage{lipsum}
\usepackage{latexsym}
\usepackage{graphicx}
\usepackage{caption}
\usepackage{blindtext}
\usepackage{multicol}
\usepackage{balance}
\usepackage{colortbl}
\usepackage{epstopdf}
\usepackage{float}
\usepackage{booktabs}
\usepackage{adjustbox}
\usepackage{tabularx}
\usepackage{makecell}
\usepackage{listings}
\usepackage{matlab-prettifier}
\usepackage{xcolor}
\usepackage{amsmath}
\usepackage{amssymb}
\usepackage{subcaption}
\usepackage{algorithm}
\usepackage{algpseudocode}
\usepackage{amsmath}
\usepackage{caption}
\usepackage{cuted}
\definecolor{TUDelftMidBlue}{HTML}{0095bf}
\usepackage[colorlinks=true, allcolors=TUDelftMidBlue]{hyperref}

\usepackage{tikz}
    \usetikzlibrary{shapes,arrows}
    \usetikzlibrary{arrows.meta}
    \usetikzlibrary{calc}
    \usetikzlibrary{intersections}
\usepackage{verbatim}
    \tikzstyle{block} = [draw, rectangle, minimum height=0.8cm, minimum width=1.3cm]
    \tikzstyle{sum} = [draw, circle] 
    \tikzstyle{input} = [coordinate]
    \tikzstyle{output} = [coordinate]
    \tikzstyle{tmp} = [coordinate]
    \tikzstyle{pinput} = [pin edge={<-,thin,black}]
    \tikzstyle{poutput} = [pin edge={->,thin,black}]

\usepackage{soul}

\graphicspath{{./Figures/}}

\title{Consensus-based Recursive \\ Multi-Output Gaussian Process}

\author{Yogesh Prasanna Kumar Rao, Tamas Keviczky, Raj Thilak Rajan 
\thanks{This work is partially funded by the EU-HORIZON-KDT-JU-2023-2-RIA, under grant agreement No 101139996, the ShapeFuture project.}
}

\begin{document}

\newgeometry{left=1.57cm,right=1.57cm,top=1.54cm,bottom=2cm}

\maketitle







\begin{abstract}
Multi-output Gaussian Processes provide principled uncertainty-aware learning of vector-valued fields but are difficult to deploy in large-scale, distributed, and streaming settings due to their computational and centralized nature. This paper proposes a Consensus-based Recursive Multi-Output Gaussian Process (CRMGP) framework that combines recursive inference on shared basis vectors with neighbour-to-neighbour information-consensus updates. The resulting method supports parallel, fully distributed learning with bounded per-step computation while preserving inter-output correlations and calibrated uncertainty. Experiments on synthetic wind fields and real LiDAR data demonstrate that CRMGP achieves competitive predictive performance and reliable uncertainty calibration, offering a scalable alternative to centralized Gaussian process models for multi-agent sensing applications.
\end{abstract}


\section{Introduction }\label{sec:introduction}

Gaussian Processes (GPs) \cite{Rasmussen2006} and their multi-output extensions (MOGPs) \cite{Liu_2018} provide a principled, non-parametric Bayesian framework for reconstructing unknown functions from sparse, noisy measurements while delivering calibrated uncertainty—properties that are highly desirable for sensing, estimation, and decision-making in robotic systems. MOGPs extend this capability to vector-valued fields (e.g., multi-component wind, temperature and humidity), explicitly modelling cross-output correlations so information can be shared across related tasks and improve reconstruction in data-scarce regions. This capability has proven useful in wind mapping, collaborative UAV mapping, and environmental monitoring where measurements are expensive or distributed \cite{8467518, en17163895, LOPEZLOPERA2022108139}.However, the conventional MOGP’s cubic cost with respect to data prevents direct use in large-scale, streaming, or resource-constrained multi-agent settings. At the same time, real-world deployments demand three properties simultaneously: (i) online, bounded-cost updates for streaming data; (ii) decentralised, neighbor-to-neighbor operation to avoid costly central aggregation; and (iii) retention of cross-output statistical coupling so multi-output benefits are preserved. This bottleneck is acute in multi-agent systems: as the volume of sensory data grows, centralized processing becomes both computationally and communicatively impractical. 

Distributed Gaussian processes decentralize computation by having each agent or expert maintain a local model and exchange only compact summaries (e.g., means, variances) rather than raw data, which reduces communication, scales efficiently, and avoids single points of failure, though standard schemes like BCM/rBCM are usually single-output and ignore cross-output correlations \cite{deisenroth2015distributed, zhai2023distributed}. Recursive Gaussian processes complement this by maintaining a compact set of inducing or basis points and updating posterior statistics incrementally with bounded per-step cost, thus enabling true streaming inference on resource-limited agents while keeping the updates exact and computationally efficient \cite{Sparse_Online_GPs}.

However, limited work has been performed to combine the benefits of both recursive GP and Distributed GP. To this end, in this work we propose Consensus-based Recursive Multi-Output Gaussian Processes (CRMGP), a framework that combines compact, recursive updates with distributed information fusion and latent-mixing structures that maintain cross-output correlations and calibrated uncertainty. CRMGPs enable each agent to perform bounded per-step inference, exchange only compact summaries with neighbors, and jointly reconstruct vector fields suitable for real-time, uncertainty-aware control and planning in multi-robot systems. The resulting approach delivers both the scalability required for large networks and the probabilistic fidelity necessary for safe, informed decision making in dynamic environments.

The remainder of the paper is organized as follows. Section II presents the preliminaries, followed by the proposed framework in Section III. Section IV evaluates the framework through simulations and compares its performance against centralized single-output and multi-output Gaussian Process models, as well as their sparse variants. Finally, Section V concludes the paper and outlines directions for future work. 
 
\section{Preliminaries}

\subsection{Multi-Output Gaussian Processes} Consider a multi-output training set 
$\mathcal{D}=\{(\mathbf{x}_i,\mathbf{y}_i)\}_{i=1}^N$, 
where $\mathbf{x}_i\in\mathbb{R}^d$ is an input and 
$\mathbf{y}_i\in\mathbb{R}^D$ is a $D$-dimensional observation. We model the vector-valued function 
$\mathbf{f}:\mathbb{R}^d\rightarrow\mathbb{R}^D$
with a multi-output Gaussian process prior and assume additive Gaussian noise
\begin{align}
\label{eq:mogp_model}
\mathbf{y}_i = \mathbf{f}(\mathbf{x}_i) + \boldsymbol{\varepsilon}_i, 
\qquad 
\boldsymbol{\varepsilon}_i \sim \mathcal{N}(\mathbf{0}, \sigma_n^2 \mathbf{I}_D).
\end{align} 
Stacking inputs and observations gives $\mathbf{X} = [\mathbf{x}_1,\dots,\mathbf{x}_N]^\top\in\mathbb{R}^{N \times d}$, $\mathbf{y}=[\mathbf{y}_1^\top,\dots,\mathbf{y}_N^\top]^\top\in\mathbb{R}^{ND}$. Let $\mathbf{K}(\mathbf{X},\mathbf{X}') \in \mathbb{R}^{ND \times MD}$ denote the block covariance matrix between two input sets 
$\mathbf{X} = [\mathbf{x}_i]_{i=1}^N$ and 
$\mathbf{X}' = [\mathbf{x}'_j]_{j=1}^M$, defined as
\begin{align}
\mathbf{K}(\mathbf{X}, \mathbf{X}')
=
\begin{bmatrix}
\mathbf{K}(\mathbf{x}_1,\mathbf{x}'_1) & \cdots & \mathbf{K}(\mathbf{x}_1,\mathbf{x}'_M) \\
\vdots & \ddots & \vdots \\
\mathbf{K}(\mathbf{x}_N,\mathbf{x}'_1) & \cdots & \mathbf{K}(\mathbf{x}_N,\mathbf{x}'_M)
\end{bmatrix},
\end{align}
which has a block structure with $D \times D$ blocks, where the $(i,j)$-th block is given by $\mathbf{K}(\mathbf{x}_i,\mathbf{x}'_j)\in\mathbb{R}^{D\times D}$ with entries $\big[\mathbf{K}(\mathbf{x}_i,\mathbf{x}'_j)\big]_{ab}
= \mathrm{Cov}\big(f_a(\mathbf{x}_i), f_b(\mathbf{x}'_j)\big).$  The moments of the joint predictive posterior at test inputs $\mathbf{X}_\ast = [\mathbf{x}_{1\ast},\dots,\mathbf{x}_{p\ast}]^\top \in\mathbb{R}^{p \times d}$ then take the form
\begin{align}
\boldsymbol{\mu}_\ast 
&= \mathbf{K}(\mathbf{X}_\ast,\mathbf{X})\,
\mathbf{K}_Y^{-1}\mathbf{y}, \\
\boldsymbol{\Sigma}_\ast
&= \mathbf{K}(\mathbf{X}_\ast,\mathbf{X}_\ast)
- \mathbf{K}(\mathbf{X}_\ast,\mathbf{X})\,
\mathbf{K}_Y^{-1}\,
\mathbf{K}(\mathbf{X},\mathbf{X}_\ast),
\end{align}
where $\mathbf{K}_Y \triangleq \mathbf{K}(\mathbf{X},\mathbf{X}) + \sigma_n^2 \mathbf{I}_{ND}$ is the full multi-output covariance. Although these expressions mirror the single-output Gaussian process, the cost is substantially higher due to larger output dimensionality $D$.

The Linear Model of Coregionalization (LMC) is a flexible way to construct valid matrix-valued kernels by expressing each output as a linear combination of $Q$ latent scalar GPs:
\begin{align}
f_d(\mathbf{x}) = \sum_{q=1}^Q a_{dq}\,u_q(\mathbf{x}),
\qquad
u_q(\cdot)\sim\mathcal{GP}(0,k_q(\cdot,\cdot)),
\end{align}
where $\mathbf{a}_q=[a_{1q},\ldots,a_{Dq}]^\top\in\mathbb{R}^D$ are coregionalization vectors. This yields the kernel
$\mathbf{K}(\mathbf{x}_1,\mathbf{x}_2)
= \sum_{q=1}^Q k_q(\mathbf{x}_1,\mathbf{x}_2)\,\mathbf{a}_q \mathbf{a}_q^\top$, so inter-output correlations are encoded by the (typically low-rank) matrices $\mathbf{a}_q\mathbf{a}_q^\top$ \cite{alvarez2010efficient, alvarez2011computationally}. Sparsity-promoting inducing-variable or variational approximations (e.g., \cite{titsias2009variational, snelson2005sparse}) can be combined with LMC to reduce the dominant cost from $\mathcal{O}((ND)^3)$ to operations on $M\times M$ matrices (with low-rank corrections), making multi-output GP inference tractable for larger problems. Alternatively, state-space or Kalman filtering formulations provides a way to achieve bounded per-step computation in streaming settings.

\subsection{Recursive Multi-Output Gaussian Processes} A Recursive Multi-Output Gaussian Process (RMGP) performs sequential Bayesian updates of the posterior distribution over the latent function, $p(\mathbf{f} \mid \mathcal{D}_{1:t})$, as new observations arrive. In principle, recursive Gaussian process inference does not require a finite set of basis inputs, although exact formulations are typically restricted to kernels admitting state-space representations. Exact recursive formulations can be obtained by exploiting the equivalence between Gaussian processes and stochastic differential equations, which allows GP regression to be cast as a (possibly infinite-dimensional) state-space model and solved via Kalman filtering \cite{e91d4272de4e442399a254c4a962ac70} or using Kriging \cite{9381509}, but is restrictive in the class of kernels that it admits.

Rather than recomputing the full posterior from scratch, RMGP maintains a compressed representation of the posterior over a fixed set of basis inputs $\mathbf{X}_b = [\mathbf{x}_1, \ldots, \mathbf{x}_M]$, where $M \ll N$ \cite{Huber2014,8571257}. Specifically, the method tracks the posterior over the inducing variables $\mathbf{g} \equiv \mathbf{f}(\mathbf{X}_b) \in \mathbb{R}^{MD}$, i.e., $p(\mathbf{g} \mid \mathcal{D}_{1:t})$, parameterized by its mean $\boldsymbol{\mu}_t^{\mathbf{g}}$ and covariance $\mathbf{C}_t^{\mathbf{g}}$. Upon receiving a new observation $(\mathbf{x}_t, \mathbf{y}_t)$, the posterior is updated recursively via a prediction step followed by a Bayesian correction, closely analogous to a Kalman filter. Consider the matrix-valued kernel $\mathbf{K}(\cdot, \cdot)$ and define
\begin{align}
\mathbf{J}_t &= \mathbf{K}(\mathbf{x}_t, \mathbf{X}_b)\mathbf{K}(\mathbf{X}_b, \mathbf{X}_b)^{-1} \in \mathbb{R}^{D \times MD},
\end{align} then the predictive distribution of the latent function value $\mathbf{f}(\mathbf{x}_t)$ has mean and covariance
\begin{align}
\boldsymbol{\mu}_t^{\mathbf{p}} &= \mathbf{m}(\mathbf{x}_t) + \mathbf{J}_t\big(\boldsymbol{\mu}_{t-1}^{\mathbf{g}} - \mathbf{m}(\mathbf{X}_b)\big), \\
\mathbf{C}_t^{\mathbf{p}} &= \mathbf{K}(\mathbf{x}_t, \mathbf{x}_t) - \mathbf{J}_t \mathbf{K}(\mathbf{X}_b, \mathbf{x}_t) + \mathbf{J}_t \mathbf{C}_{t-1}^{\mathbf{g}} \mathbf{J}_t^\top,
\end{align}
which captures the uncertainty in the latent function prior to incorporating observation noise. The update step then applies a Kalman-like correction:
\begin{align}
\mathbf{G}_t &= \mathbf{C}_{t-1}^{\mathbf{g}} \mathbf{J}_t^\top \big(\mathbf{C}_t^{\mathbf{p}} + \sigma_n^2 \mathbf{I}_D\big)^{-1}, \\
\boldsymbol{\mu}_t^{\mathbf{g}} &= \boldsymbol{\mu}_{t-1}^{\mathbf{g}} + \mathbf{G}_t \big(\mathbf{y}_t - \boldsymbol{\mu}_t^{\mathbf{p}}\big), \\
\mathbf{C}_t^{\mathbf{g}} &= \mathbf{C}_{t-1}^{\mathbf{g}} - \mathbf{G}_t \big(\mathbf{C}_t^{\mathbf{p}} + \sigma_n^2 \mathbf{I}_D\big)\mathbf{G}_t^\top.
\end{align}

This recursion implements sequential Bayesian updating while approximating the full posterior through its marginal over the inducing variables $p(\mathbf{g} \mid \mathcal{D}_{1:t})$. The procedure is initialized with $\boldsymbol{\mu}_0^{\mathbf{g}} = \mathbf{m}(\mathbf{X}_b)$ and $\mathbf{C}_0^{\mathbf{g}} = \mathbf{K}(\mathbf{X}_b, \mathbf{X}_b)$. This formulation yields bounded per-step computational complexity comparable to pseudo-input sparse Gaussian process methods, while enabling true streaming updates and immediate uncertainty-aware predictions.

The method in \cite{8571257} extends recursion to the multi-output case and studies online hyperparameter adaptation. The decentralization, however, relies on a central aggregator. The next section develops a fully distributed alternative: a Consensus Recursive Multi-Output Gaussian Process (CRMGP) that preserves exact basis-posteriors while relying solely on neighbour-to-neighbour communications to estimate the posterior mean and covariance at the basis vectors.

\section{Proposed Framework} \label{sec:proposed_framework}
We assume each agent $i\in\mathcal{V}$ maintains a local MOGP over the vector output components using the LMC prior, and updates it recursively from its own measurements, and exchanges compact information-form summaries with neighbours over the communication graph $\mathcal{G}=(\mathcal{V},\mathcal{E})$ undirected and connected, an assumption that need not necessarily hold \cite{1431045}. Furthermore, the hyperparameters are fixed and shared across nodes, moments of the local models are then fused via neighbor-to-neighbor averaging. Specifically, the natural parameters of the local Gaussian approximations are averaged using Metropolis weights, which are row-stochastic and guarantee consensus among the nodes\cite{PeterDeisenrothMDEISENROTHDistributedProcesses}. 
Distributed averaging ensures convergence under standard connectivity assumptions, including static connected graphs and time-varying graphs satisfying joint connectivity (i.e., the union of graphs over bounded time intervals is connected) \cite{xiao2007distributed}. At global time $t$, node $i$ has access to a (possibly asynchronously updated) local dataset $\mathcal{D}_{i,1:t}=\{(\mathbf{x}^i_\tau,\mathbf{y}^i_\tau)\}_{\tau=1}^{N_i(t)},\qquad
\mathbf{y}^i_\tau\in\mathbb{R}^{D},$ where $N_i(t)$ denotes the number of observations received by node $i$ up to time $t$. Nodes process only local data and exchange information parameters during consensus rounds. The global dataset is conceptually 
$\mathcal{D}_{1:t}=\biguplus_{i\in\mathcal{V}}\mathcal{D}_{i,1:t}$. Now, let the common basis vectors across all the nodes be $\mathbf{X}_b=\big[\mathbf{x}_1,\dots,\mathbf{x}_M\big] \in \mathbb{R}^{d \times M}$. For a new measurement pair $(\mathbf{x}_{i,t},\mathbf{y}_{i,t})$, we define
\begin{align}
\mathbf{J}_{i,t} &= \mathbf{K}(\mathbf{x}_{i,t},\mathbf{X}_b)\,\mathbf{K}(\mathbf{X}_b,\mathbf{X}_b)^{-1}
\in \mathbb{R}^{D \times DM}, \label{eq:J_local_short} \\
\mathbf{S}_{i,t} &\equiv \mathbf{K}(\mathbf{x}_{i,t},\mathbf{x}_{i,t})-\mathbf{J}_{i,t}\mathbf{K}(\mathbf{X}_b,\mathbf{x}_{i,t}) +\sigma_n^2\mathbf{I}_D,
\label{eq:S_local_short} 
\end{align} where $\mathbf{J}_{i,t} \in \mathbb{R}^{D \times DM}$ is the cross-covariance between the local observation and the stacked inducing variables, and $\mathbf{S}_{i,t} \in \mathbb{R}^{D \times D}$ is the effective measurement covariance including the observation noise.

\subsection{Consensus-based Recursive Gaussian Process}

Each node $i$ initializes its local information parameters for the stacked inducing variables $\mathbf{g}$ as
\begin{align}
\boldsymbol{\xi}^{\mathbf g}_{i,0} &= \mathbf{\Omega}^{\mathbf g}_{i,0} \boldsymbol{\mu}^{\mathbf g}_0, &
\mathbf{\Omega}^{\mathbf g}_{i,0} &= (\mathbf{C}^{\mathbf g}_0)^{-1},
\end{align}
where  $\boldsymbol{\mu}^{\mathbf g}_0(\mathbf{X}_b) \in \mathbb{R}^{DM}$(assumed to be zero),$ \mathbf{C}^{\mathbf g}_0 \equiv \mathbf{C}^{\mathbf g}_0(\mathbf{X}_b,\mathbf{X}_b)=\mathbf{K}(\mathbf{X}_b,\mathbf{X}_b) \in \mathbb{R}^{DM \times DM}$. Given a new measurement $(\mathbf{x}_{i,t}, \mathbf{y}_{i,t})$, the natural parameters immediately before the update are
\begin{align}
\boldsymbol{\xi}^{\mathbf g}_{i,t-1} &= \mathbf{\Omega}^{\mathbf g}_{i,t-1} \boldsymbol{\mu}^{\mathbf g}_{i,t-1}, &
\mathbf{\Omega}^{\mathbf g}_{i,t-1} &= (\mathbf{C}^{\mathbf g}_{i,t-1})^{-1}.
\end{align} and the local single-datum additive update in information form is \begin{align}
\boldsymbol{\xi}^{\mathbf g}_{i,t} &= \boldsymbol{\xi}^{\mathbf g}_{i,t-1} + \mathbf{J}_{i,t}^\top \mathbf{S}_{i,t}^{-1} \mathbf{y}_{i,t} \label{eq: info_update_short_mean}, \\
\mathbf{\Omega}^{\mathbf g}_{i,t} &= \mathbf{\Omega}^{\mathbf g}_{i,t-1} + \mathbf{J}_{i,t}^\top \mathbf{S}_{i,t}^{-1} \mathbf{J}_{i,t}.
\label{eq:info_update_short_cov}
\end{align}

Following the local update, each node performs a consensus step by averaging its information parameters with those of its neighbors $\mathcal{N}_i$ in the communication graph $\mathcal{G}$
\begin{align}
\boldsymbol{\xi}^{\mathbf{g}(l+1)}_{i,t} 
= \sum_{j \in \mathcal{\bar{N}}_i} w_{ij} \, \boldsymbol{\xi}^{\mathbf{g}(l)}_{j,t},\quad 
\mathbf{\Omega}^{\mathbf{g}(l+1)}_{i,t} 
= \sum_{j \in \mathcal{\bar{N}}_i} w_{ij} \, \mathbf{\Omega}^{\mathbf{g}(l)}_{j,t},
\label{eq:consensus_moments}
\end{align}
where $\mathcal{\bar{N}}_i = \mathcal{N}_i \cup \{i\}$, $l$ indexes the consensus iteration and $w_{ij} \ge 0$ are row-stochastic weights on $\mathcal{G}$ (e.g., Metropolis weights \cite{xiao2007distributed}). Other standard choices in the literature include fastest-mixing Markov chain weights \cite{boyd2004fastest} or optimal convex-optimization weights \cite{roch2005bounding}. The consensus step propagates information across nodes while keeping the per-node computational cost bounded.


Let $N_a = |\mathcal{V}|$ is the number of agents, then after the consensus process has sufficiently converged i.e, within a prescribed tolerance or after $L$ iterations, then each node forms an approximation of the global information by scaling its averaged increments
\begin{align}
\bar{\boldsymbol{\xi}}^{\mathbf g}_{i,T} = 
N_a\,\boldsymbol{\xi}^{\mathbf{g}(L)}_{i,T},\quad
\bar{\mathbf{\Omega}}^{\mathbf{g}}_{,i,T} 
= \mathbf{\Omega}^{\mathbf{g}(L)}_{i,0} + N_a\big(\mathbf{\Omega}^{\mathbf{ g}(L)}_{i,T}-\mathbf{\Omega}^{\mathbf{g}(L)}_{i,0}\big), \nonumber
\end{align} and recovers the basis posterior
\begin{align}
\boldsymbol{\mu}^{\mathbf g}_{i}=
\mathbf{C}^{\mathbf g}_{i}\,\bar{\boldsymbol{\xi}}^{\mathbf g}_{\text{global},i,T},\qquad
\mathbf{C}^{\mathbf g}_{i}=\big(\bar{\mathbf{\Omega}}^{\mathbf g}_{\text{global},i,T}\big)^{-1}.
\label{eq:recover_posterior_moments}
\end{align}
Predictions for any test input at node $i$ can then be made using the standard recursive form inference. The advantage here is the fact that inference can happen in parallel - leading to faster learning and exploring of the spatial environments. The proposed algorithm is summarized in Algorithm \autoref{alg:drmogp_zero_mean}.

\subsubsection*{Computational Complexity} Consider the output to be of $D$ dimension and $M$ be the number of basis vectors per node $i$. $L$ consensus rounds take place at every time-step $t$. Considering sequential arrival of data points, the per timestep computational complexity at every node is $\mathcal{O}(\text{max}(D^3, D^2M^2))$ per data point per node for the local update. The consensus update costs per node can now be computed. For the $L$ consensus iterations per time step, the costs are $\mathcal{O}(L|\mathcal{N}_i| D^2M^2)$ per node. The prediction costs remain the same as the centralized framework, albeit the distributed inference can occur in parallel.

\begin{algorithm}[t]
\caption{Consensus-based Recursive MOGP (at node $i$)}
\label{alg:drmogp_zero_mean}
\begin{algorithmic}[1]
\Require Graph $\mathcal{G}$, Shared basis $\mathbf{X}_b$, prior information $\boldsymbol{\xi}^{\mathbf{g}}_{\mathrm{prior}},\mathbf{\Omega}^{\mathbf{g}}_{\mathrm{prior}}$; consensus rounds $L$
\State \textbf{Initialize:} $\mathbf{\Omega}^{\mathbf{g}}_{i,0}\gets \mathbf{\Omega}^{\mathbf{g}}_{\mathrm{prior}}$, $\boldsymbol{\xi}^{\mathbf{g}}_{i,0}\gets \boldsymbol{\xi}^{\mathbf{g}}_{\mathrm{prior}}$ 
\For{each time step $t=1,2,\dots$}
    \If{new data pair $(\mathbf{x}_{i,t},\mathbf{y}_{i,t})$ arrives}
        \State Define $\mathbf{J}_{i,t}$ (\ref{eq:J_local_short}),  $\mathbf{S}_{i,t}$ (\ref{eq:S_local_short})
        \State Update $\boldsymbol{\xi}^{\mathbf{g}}_{i,t}$ (\ref{eq: info_update_short_mean}), $\mathbf{\Omega}^{\mathbf{g}}_{i,t}$(\ref{eq:info_update_short_cov})
    \Else
      \State $\boldsymbol{\xi}^{\mathbf{g}}_{i,t} \gets \boldsymbol{\xi}^{\mathbf{g}}_{i,t-1}$, \quad $\mathbf{\Omega}^{\mathbf{g}}_{i,t} \gets \mathbf{\Omega}^{\mathbf{g}}_{i,t-1}$  \Comment{No data}
    \EndIf
  \EndFor
  
  \For{$l=1$ \textbf{to} $L$} \Comment{Consensus}
        \State Update $\boldsymbol{\xi}^{\mathbf{g}(l+1)}_{i,t}$, $\mathbf{\Omega}^{\mathbf{g}(l+1)}_{i,t}$ (\ref{eq:consensus_moments})
  \EndFor
\State Recover posterior moments $\boldsymbol{\mu}^{\mathbf{g}}_{i}, \mathbf{C}^{\mathbf{g}}_{i}$ from (\ref{eq:recover_posterior_moments})
\State \textbf{Output:} $\boldsymbol{\mu}^{\mathbf{g}}_{i}, \mathbf{C}^{\mathbf{g}}_{i}$  
\end{algorithmic}
\end{algorithm}

\section{Results}
\begin{figure} [t]
    \centering
    \includegraphics[width=1\linewidth]{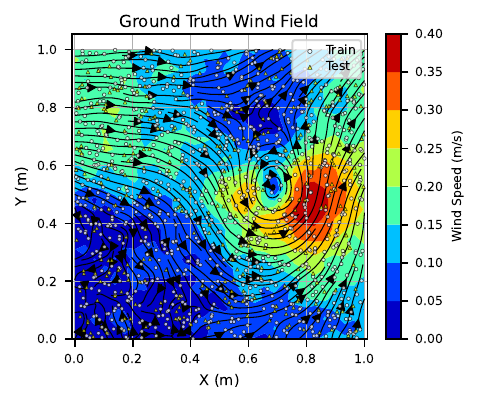}
    \caption{2D Wind Field}
    \label{fig1}
\end{figure}

\begin{figure*}[t]
  \centering
  \begin{subfigure}[b]{0.30\textwidth}
    \centering
    \includegraphics[width=\linewidth]{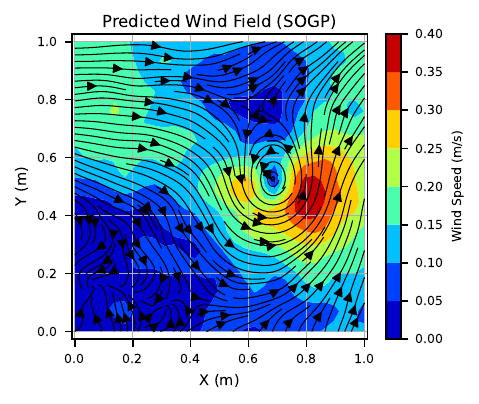}
    \caption{SOGP}
    \label{fig:1a}
  \end{subfigure}%
  \hfill
  \begin{subfigure}[b]{0.30\textwidth}
    \centering
    \includegraphics[width=\linewidth]{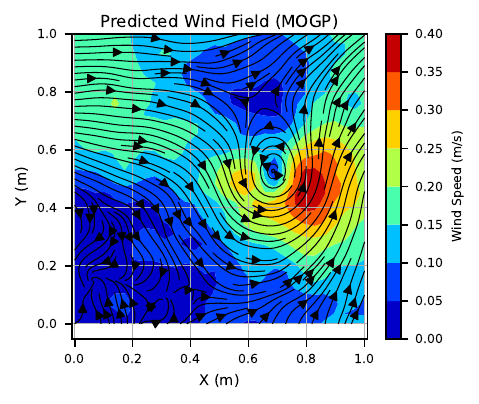}
    \caption{MOGP}
    \label{fig:1b}
  \end{subfigure}%
  \hfill
  \begin{subfigure}[b]{0.30\textwidth}
    \centering
    \includegraphics[width=\linewidth]{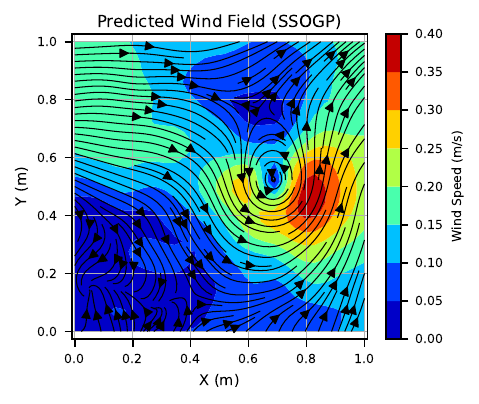}
    \caption{SSOGP}
    \label{fig:1c}
  \end{subfigure}

  \vspace{0.5em} 

  \begin{subfigure}[b]{0.30\textwidth}
    \centering
    \includegraphics[width=\linewidth]{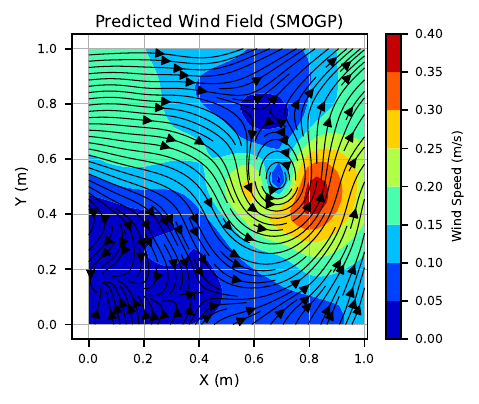}
    \caption{SMOGP}
    \label{fig:2a}
  \end{subfigure}%
  \hfill
  \begin{subfigure}[b]{0.30\textwidth}
    \centering
    \includegraphics[width=\linewidth]{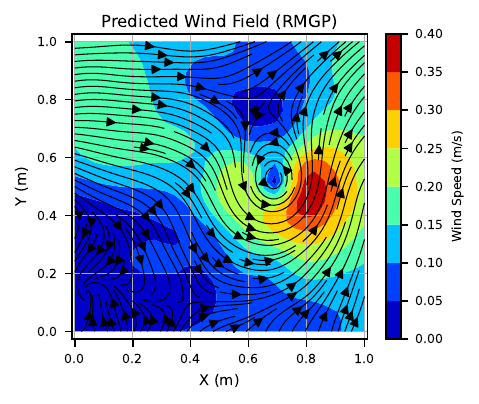}
    \caption{RMGP}
    \label{fig:2b}
  \end{subfigure}%
  \hfill
  \begin{subfigure}[b]{0.30\textwidth}
    \centering
    \includegraphics[width=\linewidth]{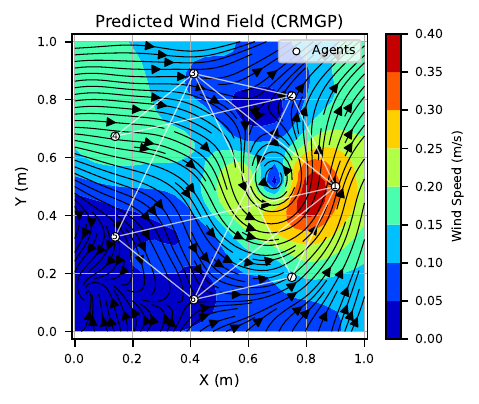}
    \caption{CRMGP}
    \label{fig:2c}
  \end{subfigure}

  \caption{Reconstruction of the wind field by each model}
  \label{fig:grid6}
\end{figure*}

\begin{figure*}[!htbp]
  \centering
  \begin{subfigure}[b]{0.30\textwidth}
    \centering
    \includegraphics[width=\linewidth]{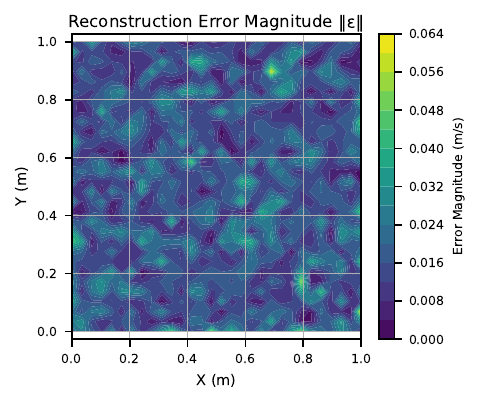}
    \caption{SOGP}
    \label{fig:1a}
  \end{subfigure}%
  \hfill
  \begin{subfigure}[b]{0.30\textwidth}
    \centering
    \includegraphics[width=\linewidth]{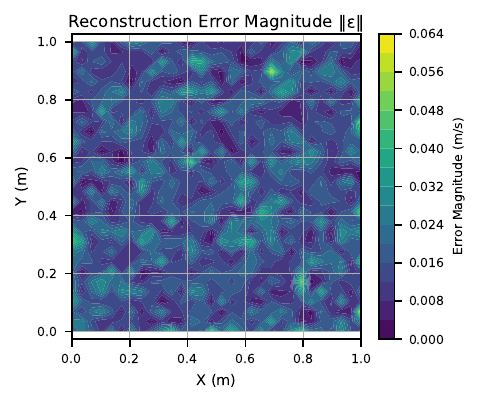}
    \caption{MOGP}
    \label{fig:1b}
  \end{subfigure}%
  \hfill
  \begin{subfigure}[b]{0.30\textwidth}
    \centering
    \includegraphics[width=\linewidth]{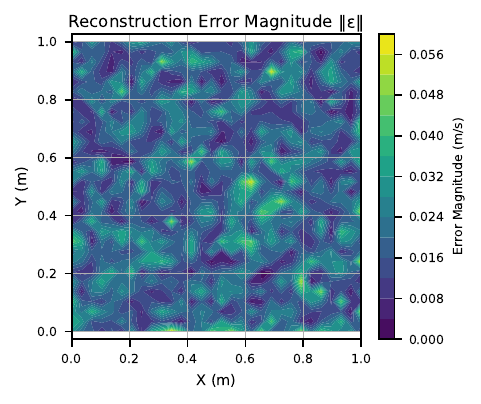}
    \caption{SSOGP}
    \label{fig:1c}
  \end{subfigure}

  \vspace{0.5em} 

  \begin{subfigure}[b]{0.30\textwidth}
    \centering
    \includegraphics[width=\linewidth]{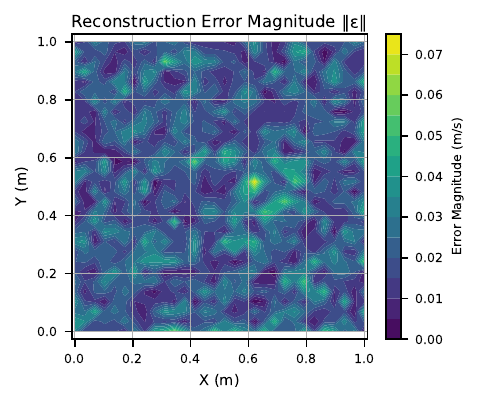}
    \caption{SMOGP}
    \label{fig:2a}
  \end{subfigure}%
  \hfill
  \begin{subfigure}[b]{0.30\textwidth}
    \centering
    \includegraphics[width=\linewidth]{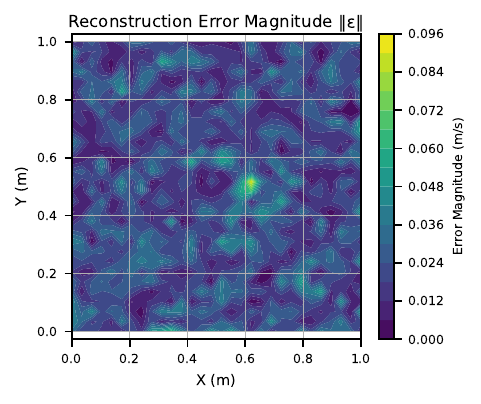}
    \caption{RMGP}
    \label{fig:2b}
  \end{subfigure}%
  \hfill
  \begin{subfigure}[b]{0.30\textwidth}
    \centering
    \includegraphics[width=\linewidth]{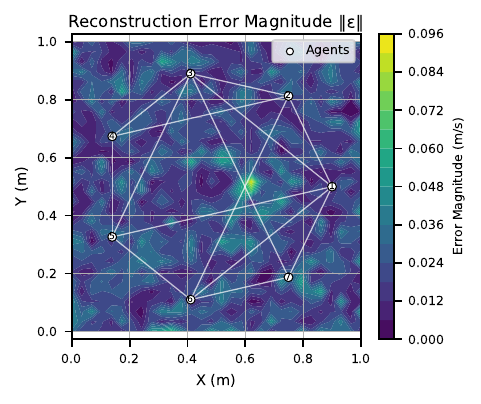}
    \caption{CRMGP}
    \label{fig:2c}
  \end{subfigure}

  \caption{Reconstruction errors of the wind field by each model}
  \label{fig:grid7}
\end{figure*}

Our proposed framework was evaluated against a synthetic 2D Wind Field shown in Figure \ref{fig1} with wind velocities (U and V) as the output in the X-Y plane. The wind field simulates wake effects experienced by downstream wind turbines. Dataset $\mathcal{D} = \{\mathbf{x}_i, \mathbf{y}_i\}^{1200}_{i = 1}$ with $N$ = 900 training points and 300 test points, where $\mathbf{x}_i \in \mathbb{R}^2$ and $\mathbf{y}_i \in \mathbb{R}^2$ were generated. The $N_a = 7$ agents were assumed to be spatially distributed along the wind field having a well-sampled dataset $\mathcal{D}_i$ having a subset of the data arbitrarily distributed.
Both single output and multi-output Gaussian Processes along with their sparse variants based on the work of \cite{titsias2009variational} were used for evaluation. Further, the multi-output version of the Recursive GP \cite{Huber2014} was used. All the models, assumed with zero mean prior, were trained using the Matern 3/2 Kernels with the multi-output GPs having the LMC formulation with $Q=2$. The code was implemented in python with the help of the GPy library \cite{gpy2014}. 
The main aim of the simulation experiment is to assess the feasibility of the proposed framework in relation to centralized schemes, highlighting the benefits of reduced computational complexity in sequential inference scenarios. The aim of the paper is not optimization of hyperparameters but rather exploring new avenues that make distributed inference more feasible and applicable to real-time scenarios. In the following results, the abbreviations are as follows: SOGP - Single Output GP \cite{Rasmussen2006}, MOGP - Multi-Output GP \cite{gpy2014}, SSOGP - Sparse Single Output GP \cite{titsias2009variational}, SMOGP - Sparse Multi-Output GP \cite{LOPEZLOPERA2022108139}, RMGP - Recursive Multi-Output GP \cite{8571257}. In \autoref{tab:testing_error_ci}, Negative Log Predictive Density (NLPD) per component, given by
\begin{align}
    \text{NLPD} = -\frac{1}{N} \sum_{i=1}^{N}\log p(\mathbf{y}_{i}^{*}|\mathbf{X,Y},\mathbf{x}_{i}^{*}),
\end{align}
which evaluates the full predictive distribution per output (wind speed in directions U and V), rewarding calibrated means, and sensible uncertainty. In combination with the Confidence intervals (CI), the comparison can be evaluated for the various models. It is evident that the Consensus-based Recursive GP model performs on par with the Multi-Output GP even when using only $M= 100$ basis vectors and $N_a = 7$ agents spread across the wind field. The statistical errors in including NLPD, Confidence interval (CI) and  Root Mean Square Erorr (RMSE) are tabulated in \autoref{tab:testing_error_ci}. The reconstruction of the field \autoref{fig1} using different methods are presented in \autoref{fig:grid6}, and their reconstruction errors are presented in \autoref{fig:grid7}.

\definecolor{trainCell}{HTML}{F2F7FF}
\definecolor{trainHead}{HTML}{E0ECFF}

\definecolor{testCell}{HTML}{F8F6F4}
\definecolor{testHead}{HTML}{E1E1E1}

\definecolor{reconCell}{HTML}{F9F1F4}
\definecolor{reconHead}{HTML}{ECEFF3}

\arrayrulecolor{black}        
\setlength{\arrayrulewidth}{0.6pt}
\renewcommand{\arraystretch}{1.15}

\begin{table}[t]
\centering
{%
\rowcolors{2}{testCell}{testCell}
\begin{tabular}{|c|c|c|c|c|}
\hline
\rowcolor{testHead}
\textbf{Model} & \textbf{NLPD U} & \textbf{NLPD V} & \textbf{95\% CI (U/V)} & RMSE \\
\hline
SOGP  & $-2.84$ & $-2.87$ & 91 / 97 & 0.020 \\
\hline
MOGP  & $-2.85$ & $-2.86$ & 94 / 96 & 0.020\\
\hline
SSOGP& $-2.79$ & $-2.75$ & 96 / 98 & 0.021\\
\hline
SMOGP & $-2.70$ & $-2.67$ & 97 / 99 & 0.023\\
\hline
RMGP  & $-2.70$ & $-2.65$ & 95 / 96 & 0.023\\
\hline
CRMGP & $-2.71$ & $-2.69$ & 95 / 97 & 0.023\\
\hline
\end{tabular}
\caption{\small Prediction performance and uncertainty of wind vector estimation. $U$ and $V$ denote the $95\%$ confidence intervals of the predicted wind components in the $x$- and $y$-directions, respectively, computed from the GP posterior variance.}
\label{tab:testing_error_ci}
}%
\end{table}

As observed from \autoref{tab:testing_error_ci}, Lower NLPD and RMSE values indicate better predictive performance, while CI coverage near 95\% reflects well-calibrated uncertainty estimates. Across these metrics, SOGP and MOGP demonstrate the strongest predictive accuracy, with NLPD values around -2.84 to -2.87 and the lowest RMSE of 0.020, suggesting they can most precisely predict wind vectors. SSOGP shows slightly reduced accuracy, and SMOGP, RMGP, and CRMGP present the highest NLPD and RMSE values, indicating marginally less precise predictions.

In terms of uncertainty quantification, MOGP achieves the best balance, with 95\% CI coverage closely matching the nominal 95\% level (94\% for U and 96\% for V). SOGP slightly underestimates uncertainty for the U component (91\%), which could risk overconfidence in predictions along that direction, while the other models—SSOGP, SMOGP, RMGP, and CRMGP—tend to overcover (95–99\%), indicating more conservative uncertainty estimates. Among these, SMOGP stands out for its particularly wide confidence intervals (97\% for U and 99\% for V), reflecting a cautious approach but at the cost of slightly higher prediction errors. While sparse batch methods reduce centralized computational complexity by compressing information into inducing inputs, CRMGP achieves comparable predictive density and uncertainty calibration while additionally enabling streaming assimilation and strictly local communication. 




Overall, the results indicate that CRMGP achieves centralized-level predictive quality and uncertainty calibration while satisfying the structural requirements of distributed, real-time inference. The framework therefore offers a practically meaningful balance between probabilistic fidelity and computational scalability, making it suitable for multi-robot wind estimation and other spatial disturbance reconstruction tasks where both uncertainty awareness and decentralized operation are essential.

\section{Conclusion} In this paper, we presented a Consensus-based Recursive Multi-Output Gaussian Process (CRMGP) for scalable, uncertainty-aware learning of vector-valued fields in distributed and streaming environments. By combining recursive GP updates with neighbor-to-neighbor consensus, we enable multi-output inference without centralized data or batch retraining. We demonstrated through simulations that CRMGP maintains reliable predictive uncertainty and competitive performance compared to centralized and sparse multi-output GPs. In our future work, we will explore advanced distributed optimization methods e.g., ADMM and extend the framework to online hyperparameter adaptation, adaptive basis selection, dynamic communication graphs, and deployment on multi-robot platforms for closed-loop sensing and control.

\bibliographystyle{IEEEtran}

\bibliography{References}

@inproceedings{zhai2023distributed,
  title={Distributed Gaussian process hyperparameter optimization for multi-agent systems},
  author={Zhai, Peiyuan and Rajan, Raj Thilak},
  booktitle={International Conference on Acoustics, Speech and Signal Processing (ICASSP)},
  pages={1--5},
  year={2023},
  organization={IEEE}
}

@article{Huber2014,
   author = {Marco F. Huber},
   doi = {10.1016/j.patrec.2014.03.004},
   issn = {01678655},
   issue = {1},
   journal = {Pattern Recognition Letters},
   month = {8},
   pages = {85-91},
   publisher = {Elsevier},
   title = {Recursive Gaussian process: On-line regression and learning},
   volume = {45},
   year = {2014}
}

@book{Rasmussen2006,
   title = {Gaussian processes for machine learning},
   author = {Carl Edward. Rasmussen and Christopher K. I.. Williams},
   isbn = {026218253X},
   pages = {248},
   publisher = {MIT Press},
   year = {2006}
}

@article{Liu_2018,
  title={Remarks on multi-output Gaussian process regression},
  author={Liu, Haitao and Cai, Jianfei and Ong, Yew-Soon},
  journal={Knowledge-Based Systems},
  volume={144},
  pages={102--121},
  year={2018},
  publisher={Elsevier}
}

@inproceedings{titsias2009variational,
  title="{Variational learning of inducing variables in sparse Gaussian processes}",
  author={Titsias, Michalis},
  booktitle={Artificial intelligence and statistics},
  pages={567--574},
  year={2009},
  organization={PMLR}
}

@inproceedings{PeterDeisenrothMDEISENROTHDistributedProcesses,
  title={Distributed gaussian processes},
  author={Deisenroth, Marc and Ng, Jun Wei},
  booktitle={International conference on machine learning},
  pages={1481--1490},
  year={2015},
  organization={PMLR}
}

@article{alvarez2011computationally,
  title={Computationally efficient convolved multiple output Gaussian processes},
  author={Alvarez, Mauricio A and Lawrence, Neil D},
  journal={The Journal of Machine Learning Research},
  volume={12},
  pages={1459--1500},
  year={2011},
  publisher={JMLR. org}
}

@inproceedings{alvarez2010efficient,
  title={Efficient multioutput Gaussian processes through variational inducing kernels},
  author={{\'A}lvarez, Mauricio and Luengo, David and Titsias, Michalis and Lawrence, Neil D},
  booktitle={Proceedings of the Thirteenth International Conference on Artificial Intelligence and Statistics},
  pages={25--32},
  year={2010},
  organization={JMLR Workshop and Conference Proceedings}
}

@article{snelson2005sparse,
  title="{Sparse Gaussian processes using pseudo-inputs}",
  author={Snelson, Edward and Ghahramani, Zoubin},
  journal={Advances in neural information processing systems},
  volume={18},
  year={2005}
}

@ARTICLE{8571257,
  author={Yang, Le and Wang, Ke and Mihaylova, Lyudmila},
  journal={IEEE Transactions on Signal and Information Processing over Networks}, 
  title={Online Sparse Multi-Output Gaussian Process Regression and Learning}, 
  year={2019},
  volume={5},
  number={2},
  pages={258-272},
  doi={10.1109/TSIPN.2018.2885925}}

@article{xiao2007distributed,
  title={Distributed average consensus with least-mean-square deviation},
  author={Xiao, Lin and Boyd, Stephen and Kim, Seung-Jean},
  journal={Journal of parallel and distributed computing},
  volume={67},
  number={1},
  pages={33--46},
  year={2007},
  publisher={Elsevier}
}

@Misc{gpy2014,
  author =   {{GPy}},
  title =    {{GPy: A Gaussian process framework in python}},
  howpublished = {\url{http://github.com/SheffieldML/GPy}},
  year = {since 2012}
}

@article{Sparse_Online_GPs,
    author = {Csató, Lehel and Opper, Manfred},
    title = {Sparse On-Line Gaussian Processes},
    journal = {Neural Computation},
    volume = {14},
    number = {3},
    pages = {641-668},
    year = {2002},
    month = {03},
    issn = {0899-7667},
    eprint = {https://direct.mit.edu/neco/article-pdf/14/3/641/815172/089976602317250933.pdf},
}

@inproceedings{deisenroth2015distributed,
  title={Distributed gaussian processes},
  author={Deisenroth, Marc and Ng, Jun Wei},
  booktitle={International conference on machine learning},
  pages={1481--1490},
  year={2015},
  organization={PMLR}
}

@article{8467518,
  author={Liu, Miao and Chowdhary, Girish and Castra da Silva, Bruno and Liu, Shih-Yuan and How, Jonathan P.},
  journal={IEEE Control Systems Magazine}, 
  title={Gaussian Processes for Learning and Control: A Tutorial with Examples}, 
  year={2018},
  volume={38},
  number={5},
  pages={53-86},
  doi={10.1109/MCS.2018.2851010}}

@article{LOPEZLOPERA2022108139,
title = {Multioutput Gaussian processes with functional data: A study on coastal flood hazard assessment},
journal = {Reliability Engineering \& System Safety},
volume = {218},
pages = {108139},
year = {2022},
issn = {0951-8320},
author = {Andrés F. López-Lopera and Déborah Idier and Jérémy Rohmer and François Bachoc}
}

@Article{en17163895,
AUTHOR = {Foley, Emma},
TITLE = {Leveraging Gaussian Processes in Remote Sensing},
JOURNAL = {Energies},
VOLUME = {17},
YEAR = {2024},
NUMBER = {16},
ARTICLE-NUMBER = {3895},
ISSN = {1996-1073},
DOI = {10.3390/en17163895}
}

@article{boyd2004fastest,
  title   = {Fastest Mixing Markov Chain on a Graph},
  author  = {Boyd, Stephen and Diaconis, Persi and Xiao, Lin},
  journal = {SIAM Review},
  volume  = {46},
  number  = {4},
  pages   = {667--689},
  year    = {2004},
  doi     = {10.1137/S0036144503423264}
}

@article{roch2005bounding,
  title   = {Bounding Fastest Mixing},
  author  = {Roch, S{\'e}bastien},
  journal = {Electronic Communications in Probability},
  volume  = {10},
  pages   = {282--296},
  year    = {2005},
  doi     = {10.1214/ECP.v10-1169}
}

@inproceedings{e91d4272de4e442399a254c4a962ac70,
title = "Kalman Filtering and Smoothing Solutions to Temporal Gaussian Process Regression Models",
author = "Jouni Hartikainen and Simo S{\"a}rkk{\"a}",
year = "2010",
language = "English",
booktitle = "MLSP IEEE International Workshop on Machine Learning for Signal Processing (MLSP), Kittil{\"a}, 29.8-1.9.2010",
}

@ARTICLE{9381509,
  author={Zhan, Dawei and Xing, Huanlai},
  journal={IEEE Transactions on Evolutionary Computation}, 
  title={A Fast Kriging-Assisted Evolutionary Algorithm Based on Incremental Learning}, 
  year={2021},
  volume={25},
  number={5},
  pages={941-955},
  doi={10.1109/TEVC.2021.3067015}}

@ARTICLE{1431045,
  author={Wei Ren and Beard, R.W.},
  journal={IEEE Transactions on Automatic Control}, 
  title={Consensus seeking in multiagent systems under dynamically changing interaction topologies}, 
  year={2005},
  volume={50},
  number={5},
  pages={655-661},
  doi={10.1109/TAC.2005.846556},
  ISSN={1558-2523},
  month={May},}

\end{document}